\documentclass{article}

\usepackage[preprint]{cpal_2025}

\usepackage[utf8]{inputenc} 
\usepackage[hidelinks, colorlinks, citecolor=blue!70!black, linkcolor=blue!70!black]{hyperref}       
\usepackage{url}            
\usepackage{booktabs}       
\usepackage{amsfonts}       
\usepackage{nicefrac}       
\usepackage{microtype}      
\usepackage{lipsum}
\usepackage{graphicx}
\graphicspath{ {./images/} }
\usepackage{ulem}
\usepackage{balance}
\usepackage{algorithm}
\usepackage{algorithmic}
\usepackage{amsmath} 
\usepackage{amssymb} 
\usepackage{adjustbox}
\usepackage{booktabs}
\usepackage{listings}
\usepackage{multirow}
\usepackage{microtype}
\usepackage{newfloat}
\usepackage{subcaption}
\usepackage[table,xcdraw]{xcolor}
\usepackage{tcolorbox}
\usepackage{xspace} 

\title{Effective Learning for Small Reasoning Models: An Empirical Study on 0.5B Reasoning LLMs}

\author {
    Xialie~Zhuang\textsuperscript{\rm 1,3}, Peixian~Ma\textsuperscript{\rm 2}, Zhikai~Jia\textsuperscript{\rm 3},  Zane~Cao\textsuperscript{\rm 3},   Shiwei~Liu\textsuperscript{\rm 4,5,6,*}\\
    \textsuperscript{\rm 1}University of Chinese Academy of Sciences, China\\
    \textsuperscript{\rm 2}The Hong Kong University of Science and Technology (Guangzhou), China\\
    \textsuperscript{\rm 3}SCITIX (SGP) TECH PTE. LTD., Singapore\\
    \textsuperscript{\rm 4}Max Planck Institute for Intelligent Systems, Germany\\
    \textsuperscript{\rm 5}ELLIS Institute Tübingen, Germany\\
    \textsuperscript{\rm 6}Tübingen AI Center, Germany\\
    \textsuperscript{\rm *}Corresponding authors: sliu@tue.ellis.eu
}
\definecolor{cityblue}{RGB}{128, 159, 225}

\begin{document}

\maketitle

\begin{abstract}
The ongoing evolution of language models has led to the development of large-scale architectures that demonstrate exceptional performance across a wide range of tasks. However, these models come with significant computational and energy demands, as well as potential privacy implications. In this context, Small Reasoning Language Models (SRLMs) with approximately 0.5 billion parameters present a compelling alternative due to their remarkable computational efficiency and cost-effectiveness, particularly in resource-constrained environments. Despite these advantages, the limited capacity of 0.5 billion parameter models poses challenges in handling complex tasks such as mathematical reasoning. This research investigates various training strategies, including supervised fine-tuning (SFT), knowledge distillation (KD), and reinforcement learning (RL), as well as their hybrid implementations, to enhance the performance of 0.5B SRLMs. We analyze effective methodologies to bridge the performance gap between SRLMS and larger models and present insights into optimal training pipelines tailored for these smaller architectures. Through extensive experimental validation and analysis, our work aims to provide actionable recommendations for maximizing the reasoning capabilities of 0.5B models.
\end{abstract}

\section{Introduction}

The field of Language Models (LMs) has experienced a significant transformation with the advent of increasingly large models, which have demonstrated outstanding performance across a range of tasks~\cite{zhao2023survey, liu2024survey, hadi2023survey}.  However, the pursuit of such massive models has introduced considerable challenges, including substantial computational requirements, increased energy consumption, and growing privacy concerns.  In this context, Small Reasoning Language Models (SRLMs) with approximately 0.5 billion parameters have emerged as a compelling alternative~\cite{van2024survey, lu2024small}.  These models not only provide enhanced computational efficiency but also offer cost-effectiveness, making them especially suitable for deployment in resource-constrained environments.

Despite their efficiency, SRLMs face inherent limitations due to their reduced size.  The constrained capacity of 0.5 billion parameters presents a significant challenge when addressing complex tasks such as mathematical reasoning. These tasks require not only an understanding of intricate concepts but also the ability to produce precise and logical outputs. Early empirical evidence suggests that while SRLMs show potential in these areas, they exhibit a noticeable performance gap compared to their larger counterparts. This gap underscores the need for effective enhancement techniques to improve the capabilities of 0.5 billion parameter models. In recent studies, researchers have employed a variety of training strategies and pipelines to align the output of LMs with target preferences, such as supervised fine-tuning~(SFT), knowledge distillation~(KD), and reinforcement learning~(RL)~\cite{guo2025deepseek, dang2025reinforcement}. However, in specific tasks, the training strategies of SRLMs have not been clearly explored, which makes it quite challenging to explore their capability boundaries and application fields.

In addressing this gap, our research investigates a range of training strategies, including SFT, KD, RL, as well as their hybrid configurations.   Each of these methodologies presents distinct advantages and challenges, particularly regarding the enhancement of 0.5B SRLMs for mathematical reasoning.   Our objective is to offer a comprehensive analysis of these approaches and their applicability to them.

Our research is driven by several key research questions:

\textit{\textbf{RQ1}: What is the capacity of 0.5B models in reasoning tasks?} To address this inquiry, we conduct a comparison of 0.5B models with both closed-sourced and open-sourced models from 1.5B to 7B. Through an analysis of baseline performance and the optimal outcomes achieved via various training methodologies, we aim to explore the gap between SRLMs and larger LMs.    This comparative study yields insights into the potential of 0.5B models and underscores the significance of training approaches in unlocking their full capabilities.

\textit{\textbf{RQ2}: How do different training strategies, including SFT, RL, and KD, specifically enhance the performance of 0.5B models in these tasks?} In this exploration, we will meticulously examine each training method's nuances and assess their individual contributions to model performance. By isolating the effects of SFT, RL, and KD, our objective is to deliver a comprehensive understanding of how these strategies can be effectively utilized to optimize the capabilities of 0.5B models in mathematical reasoning.

\textit{\textbf{RQ3}: What is the optimal combination of SFT, RL, and KD methods for enhancing the performance of 0.5B models in these domains?} Here, we will investigate the synergistic effects of integrating various training techniques. By experimenting with diverse hybrid approaches, we aim to identify the most effective pipeline that harnesses the strengths of SFT, RL, and KD. This question seeks to reveal how the combination of these methodologies can lead to significant enhancements in the performance of 0.5B models, thereby offering a practical framework for optimizing their capabilities in mathematical reasoning tasks.

In summary, this research makes the following key contributions:
\begin{itemize}
    \item We provide a thorough analysis of the capabilities of 0.5B language models in mathematical reasoning, highlighting their potential and limitations.
    \item We systematically evaluate the effectiveness of various enhancement pipelines, including SFT, RL, and KD, and their hybrid combinations, offering valuable insights into the most effective methods for improving the performance of 0.5B models.
    \item We establish practical guidance for applying these enhancement techniques to 0.5B models, helping to maximize their potential in specialized tasks.
    \item We propose future research directions aimed at further enhancing the capabilities of SRLMs, with the goal of making advanced AI more accessible and efficient for a wider range of applications.
\end{itemize}

\section{Related Works}

\subsection{Mathematical Reasoning}

Mathematical reasoning is a complex cognitive task that requires understanding, manipulating, and solving mathematical problems \cite{Creswell2022, Drori2022, Lewkowycz2022}. For language models, tasks involving mathematical reasoning necessitate comprehending mathematical concepts, applying logical reasoning, and performing calculations to reach correct solutions \cite{Cobbe2021, Lample2022, Frieder2023}. This includes solving word problems, proving theorems, and carrying out algebraic manipulations \cite{Hendrycks2021}. The model must show an understanding of mathematical principles, follow multi-step reasoning processes, and generate accurate, well-structured answers \cite{Wang2023, Gao2022, liu2023mathematical, magister2023teaching, ahn2024large, sun2024mm, lu2023mathvista}.

\subsection{Small Reasoning Language Models}
Over the past decade, language modeling has evolved significantly, transitioning from early models limited by size and data to the era of large language models (LLMs), which leverage pre-training on extensive corpora to achieve remarkable natural language understanding and generation~\cite{zhao2023survey, liu2024survey, hadi2023survey}. LLMs, exemplified by models like GPT-3~\cite{floridi2020gpt}, possess billions of parameters and excel at tasks such as text completion, translation, and summarization, yet their immense computational demands~\cite{kaddour2023challenges} restrict accessibility for smaller organizations. In response, researchers have developed Small Reasoning Language Models (SRLMs)—typically with 0.5-1.0 billion parameters, such as TinyLlama~\cite{zhang2024tinyllama}, Qwen2.5~\cite{qwen2.5}, and Qwen3~\cite{qwen3}—which balance performance and efficiency for deployment in resource-constrained environments. Motivated by the need for edge-device compatibility~\cite{lu2024small}, privacy, and sustainability, SRLMs offer practical solutions for mobile and IoT applications~\cite{li2025efficient}, with benefits including lower latency and faster response times. Their use extends to mathematical reasoning, education, and code generation, supporting both productivity and learning. Although SRLMs achieve impressive results through advanced optimization and training, their smaller size imposes limitations on complex reasoning tasks~\cite{ranaldi2024aligning, souza2025code, sqlr1}. To address these challenges, enhancement techniques such as supervised fine-tuning, reinforcement learning~\cite{dang2025reinforcement}, and knowledge distillation~\cite{chen2025sft} have been developed, further expanding their capabilities while maintaining resource efficiency.

The objective of this research is to investigate these enhancement techniques and their applications to 0.5 billion parameter models within the specific domains of mathematical reasoning. By tackling the challenges inherent in SRLMs and capitalizing on the opportunities they present, we aspire to contribute to the advancement of efficient, specialized AI models that deliver high performance while minimizing resource requirements. 

\section{Experimental Setup}
Our experimental methodology focuses on rigorously evaluating the discussed enhancement pipelines on selected 0.5B models.


\subsection{Data Preparation}
In this study, we utilize the GSM8K training set~\cite{cobbe2021gsm8k} as a primary resource for both SFT and RL training intended to enhance the mathematical reasoning capabilities of 0.5B models.    
It comprises 7K multi-step arithmetic problems and undergoes a rigorous preprocessing regimen that includes deduplication, tokenization, and Chain of Thought (CoT) augmentation~\cite{wei2022chain}.   
This process is designed to create well-structured examples that facilitate reasoning.
In addition to advancing mathematical reasoning, this training initiative aims to explore the potential for knowledge transfer.    Specifically, we evaluate whether the logical and structured reasoning skills acquired through GSM8K can enhance the models' performance in other math reasoning.


\subsection{Foundation Models}
We investigate the performance of three widely utilized SRLMs, including the Qwen2.5-0.5B-Instruct~\cite{qwen2.5}, Qwen2.5-0.5B~\cite{qwen2.5}, to assess their efficacy across various natural language processing tasks, particularly in mathematical reasoning. The Qwen2.5-0.5B-Instruct is specifically engineered for instruction-following tasks, demonstrating significant proficiency in generating responses that closely align with user prompts. In contrast, the Qwen2.5-0.5B serves as a versatile, general-purpose model, providing a robust foundation for a multitude of applications.

\subsection{Training Strategies for Exploration}
In this section, we introduce methodologies that applied in the training process of SRLMs.

\paragraph{Supervised Fine-Tuning.}
Supervised Fine-Tuning (SFT) involves the adaptation of pre-trained models for specific tasks, particularly models with 0.5 billion parameters, with the primary objective of optimizing learning from limited resources while mitigating the risks of catastrophic forgetting.  Among various methodologies, Low-Rank Adaptation (LoRA) is notable for its efficiency, as it integrates trainable low-rank matrices into Transformer layers, significantly reducing both the number of trainable parameters and memory requirements, thereby alleviating computational burdens and addressing issues of catastrophic forgetting.  However, the effectiveness of SFT is inherently dependent on the scale and quality of labeled data, which can impose a generalization ceiling and result in surface-level mimicry.  Additionally, challenges such as distribution distortion and catastrophic forgetting remain pertinent, particularly for smaller model architectures.  This study seeks to conduct a comprehensive evaluation of both full-parameter fine-tuning and LoRA fine-tuning strategies to identify the optimal SFT approach.

\paragraph{Reinforcement Learning.}

In this study, we employ the Group Relative Policy Optimization~(GRPO) algorithm in reinforcement learning~(RL), as it mitigates the necessity for a value model, reduces memory requirements, and allows for a clear articulation of reward targets, making it an optimal choice for the effective optimization of the policy model~\cite{grpo}.    
A pivotal aspect of GRPO is the design of an effective reward function, which must provide clear and informative signals to guide the model's learning process.    
Consequently, we have developed a specific reward function tailored to enhance the model's performance in mathematical reasoning tasks.    
The formulation of the reward function is delineated as follows:

\textit{\textbf{Format Reward.}} We push the model to enclose the reasoning process within \texttt{<think>...</think>} tags and to present the final answer enclosed within \texttt{<answer>...</answer>} tags. The structure of the format reward function is delineated as follows:

\begin{align}
S_{format}= 
\begin{cases} 
1, & \text{if format is correct} \\
-1, & \text{if format is incorrect}
\end{cases}
\end{align}

\textit{\textbf{Accuracy Reward.}} The accuracy of reasoning results is key criterion in the evaluation. We prioritize the Accuracy Reward as a second component of the reward functions:

\begin{align}
S_{accuracy}= 
\begin{cases} 
1, & \text{if answer is correct} \\
-1, & \text{if answer is incorrect}
\end{cases}
\end{align}

In section~\ref{sec:detail_rl}, we give a detailed description of RL algorithm.

\paragraph{Knowledge Distillation.}
Knowledge Distillation~(KD) transfers capabilities from large teacher models to smaller student models. In our pursuit of effective knowledge distillation for SRLMs, we employ the GSM8K distillation dataset, which is generated using the CAMEL framework~\cite{li2023camel}.   This dataset comprises a collection of mathematical problem-solving traces, where each entry includes a problem statement accompanied by a comprehensive step-by-step solution.   Our adherence to the methodology outlined by the CAMEL framework ensures that the distillation data is of high quality and maintains a structured format, which is essential for the efficient transfer of knowledge from the teacher model to our smaller models.   By leveraging this specialized dataset, we aim to significantly enhance the mathematical reasoning capabilities of our models through the process of knowledge distillation.

\paragraph{Hybrid Approaches.}
Hybrid approaches that integrate SFT, RL, and KD have emerged as an effective strategy to leverage the strengths of these methodologies.   Typically, SFT is implemented first to establish a robust baseline, adapting the pre-trained model to the specific task using labeled data.   This initial phase sets the groundwork for further refinement through RL, which encourages exploration and utilizes feedback mechanisms to address the generalization limitations inherent in SFT.   Additionally, KD is employed to transfer knowledge from a larger teacher model to a smaller 0.5B student model, thereby imparting general reasoning capabilities.   This foundational knowledge can subsequently be specialized through either further SFT or RL.

We hope to explore a hybrid training pipeline paradigm suitable for SRLMs in our experiments, which outperform individual approaches by strategically balancing their strengths and weaknesses.     While SFT offers a strong foundation, RL enhances the model's adaptability and generalization.     However, the effectiveness of these combinations can be task-dependent.     For instance, in certain scenarios, such as tool-calling tasks, pure RL methods, like Generalized Policy Optimization (GRPO), have shown superior performance compared to the SFT-then-RL sequence, indicating that SFT might introduce suboptimal biases in specific contexts.     Consequently, the decision to adopt a hybrid approach should be grounded in empirical validation, ensuring that the added complexity yields significant enhancements in the model's capabilities across varied tasks.


\subsection{Evaluation Benchmark \& Metric}
The evaluation of SRLMs training is conducted across several benchmarks, including OlympiadBench~\cite{he2024olympiadbench}, MATH500~\cite{hendrycks2021measuring}, MINERVA~\cite{Minervamath}, AMC23~\cite{AMC23}, and GSM8K test set~\cite{cobbe2021gsm8k}.   OlympiadBench consists of advanced mathematical problems similar to those found in mathematical olympiads.   MATH500 challenges models with complex mathematical concepts and problem-solving techniques.   MINERVA focuses on scientific reasoning, requiring the integration of scientific knowledge with mathematical skills.   AMC23 offers a unique set of mathematical challenges, while GSM8K specifically assesses the ability to handle grade school-level arithmetic problems that involve multi-step reasoning.   Each benchmark is selected for its unique contribution to evaluating different aspects of mathematical reasoning.

Performance of each LMs will be measured using accuracy as the primary metric across all mathematical reasoning benchmarks . All evaluations are conducted using the LightEval framework~\cite{lighteval} with the following configuration: models are evaluated with a maximum model length of 32768 tokens and 95\% GPU memory utilization, using bfloat16 data type and data parallel processing across multiple GPUs. Generation parameters include maximum new tokens of 32768, temperature of 0.6, and top-p sampling of 0.95 to ensure robust and consistent evaluation results. We will also report computational costs, including training time and GPU resources utilized for each enhancement pipeline.

\section{Experimental Results}

In this section, we investigate the capability boundaries of SRLMs through a series of experiments.      
\textbf{Experiment 1} conducts a comparative analysis of existing SRLMs against various parameter-scale LMs to evaluate performance differences.      
\textbf{Experiment 2} explores the influence of distinct single training strategies on the improvement of SRLMs    
\textbf{Experiment 3} examines the effects of a hybrid training strategy, providing insights into the advantages of integrating multiple training approaches. 

\subsection{Experiment 1: Comparison and Analysis of Different LMs} 

\begin{tcolorbox}[boxrule=1pt, colback=gray!25, colframe=black]
   \textbf{Answer of \textit{RQ1}}:
    While SRLMs clearly fall short of larger LLMs in mathematical reasoning, their performance is steadily improving through sophisticated techniques.
\end{tcolorbox}

As demonstrated in Table~\ref{tab:main_comp}, there is a distinct performance disparity between SRLMs and the larger scale LMs across various mathematical reasoning benchmarks.
Specifically, the Qwen2.5-0.5B-Instruct model demonstrates a baseline performance of merely 6.2\% on OlympiadBench and 31.4\% on MATH500, significantly lagging behind the 7 billion and 1.5 billion parameter models, while the Qwen2.5-7B-Instruct model achieves an impressive 38.2\% on OlympiadBench and 75.2\% on MATH500, Qwen2.5-1.5B-Instruct model attains 16.7\% on OlympiadBench and 51.2\% on MATH500.
These results underscore the inherent limitations associated with 0.5B SRLMs, attributing their subpar performance to their relatively small parameter scale and weak reasoning ability.

The implementation of effective training strategies has led to significant enhancements in the performance of 0.5B models.   
For instance, the Qwen2.5-0.5B-Instruct model, following optimization, demonstrates accuracy of 7.6\% on OlympiadBench and 32.4\% on MATH500.   
Despite significant enhancements, the performance of 0.5B SRLMs continues to fall short compared to their larger LMs.    
However, some optimized SRLMs have demonstrated a notable reduction in this performance gap. For example, Qwen3-0.6B~\cite{qwen3} achieve performance metrics of 20.2\% on OlympiadBench and 56.4\% on MATH500 after undergoing similar training optimizations.   These results underscore that appropriate training methodologies can effectively extend the capability boundaries of SRLMs. With further refinements in training strategies and improvements in model architecture, 0.5B models have the potential to attain performance levels that are increasingly comparable to those of larger models.  
This outcome demonstrates that the true capability boundaries of SRLMs are not static; rather, they can be expanded through optimized training methodologies.

\begin{table}[!t]
    \centering
    \caption{Benchmarking LLMs at different scales. We follow the results of 7B and 1.5B models from~\cite{deepscaler2025, zeng7b}. OB represents OlympiadBench. $^{\dagger}$ represents that we can not extract answer from model's response. Dash~(-) represents unavailable official results. All results are evaluated by Lighteval framework~\cite{lighteval}.}
    \begin{tabular}{l|ccccc}
        \toprule
        \textbf{Foundation Model}  & \textbf{OB} & \textbf{MATH500} & \textbf{MINERVA} & \textbf{AMC23} & \textbf{GSM8K} \\
        \midrule
        \multicolumn{6}{c}{\cellcolor[HTML]{ECF4FF}\textbf{\textit{LMs~(7B)}}} \\
        \midrule
        Qwen2.5-7B-Instruct             & 38.2 & 75.2 & 35.2 & 55.0 & 90.6   \\ 
        Qwen2.5-Math-7B-Instruct        & 40.7 & 79.8 & 34.6 & 45.0 & -  \\ 
        Deepseek-R1-Distilled-7B        & 49.8 & 92.8 &  & 77.5 &   \\ 
        Openthinker-7B                  & - & 83.2 & - & 74.5 & - \\ 
        \midrule
        \multicolumn{6}{c}{\cellcolor[HTML]{ECF4FF}\textbf{\textit{LMs~(1.5B)}}} \\
        \midrule
        Qwen2.5-1.5B-Instruct           & - & 9.0 & - & - & 0.2 \\ 
        Qwen2.5-Math-1.5B-Instruct      & 16.7 & 51.2 & - & 22.5 & -   \\
        Still-3-1.5B-Preview            & 45.4 & 84.4 & 29.0 & 66.7 & -   \\
        Deepseek-R1-Distilled-1.5B      & 31.4 & 83.9 & 26.5 & 55.0 & -      \\ 
        \midrule
        \multicolumn{6}{c}{\cellcolor[HTML]{ECF4FF}\textbf{\textit{SRLMs~(0.5B)}}} \\
        \midrule
        Qwen2.5-0.5B-Instruct           & 6.2 & 31.4 & 6.3 & 5.0 & 45.5   \\              
        Qwen2.5-0.5B$^{\dagger}$        & - & - & - & - & -                             \\   
        Qwen3-0.6B~(No Thinking)        & 19.4 & 55.6 & 16.5 & 25.0 & 73.7   \\  
        Qwen3-0.6B~(Thinking)           & 20.2 & 56.4 & 12.8 & 27.5 & 74.5   \\    
        \midrule
        \multicolumn{6}{c}{\cellcolor[HTML]{ECF4FF}\textbf{\textit{After Efficienct Training - SRLMs~(0.5B)}}} \\
        \midrule
        Qwen2.5-0.5B-Instruct           & 7.6 & 32.4 & 8.5 & 10.0 & 54.0   \\              
        Qwen2.5-0.5B                    & 5.2 & 23.6 & 4.4 & 5.0 & 34.2                 \\   
        \bottomrule
    \end{tabular}
    \label{tab:main_comp}
\end{table}

\begin{table}[!t]
    \centering
    \caption{Comparison of different training strategies. OB represents OlympiadBench. $^{\dagger}$ represents that we can not extract answer from model's response. Dash~(-) represents unavailable official results. All results are evaluated by Lighteval framework~\cite{lighteval}.}
    \begin{tabular}{l|ccccc}
        \toprule
        \textbf{Method} & \textbf{OB} & \textbf{MATH500} & \textbf{MINERVA} & \textbf{AMC23} & \textbf{GSM8K} \\
        \midrule
        \multicolumn{6}{c}{\cellcolor[HTML]{ECF4FF}\textbf{\textit{Baselines}}} \\
        \midrule
        Qwen2.5-0.5B$^{\dagger}$        & - & - & - & - & -                             \\
        Qwen2.5-0.5B-Instruct           & \uline{6.2} & \uline{31.4} & \uline{6.3} & 5.0 & 45.5 \\  
        \midrule
        \multicolumn{6}{c}{\cellcolor[HTML]{ECF4FF}\textbf{\textit{Training on Base Model}}} \\
        \midrule
        Qwen2.5-0.5B+KD        & 3.3 & 10.0 & 1.8 & 7.5 & 18.7                 \\
        Qwen2.5-0.5B+KD~(Lora)   & 2.2 & 7.6 & 2.5 & 0.0 & 9.7                   \\
        Qwen2.5-0.5B+SFT       & 3.7 & 9.2 & 1.4 & 5.0 & 21.6                  \\
        Qwen2.5-0.5B+SFT~(Lora)  & 0.7 & 1.2 & 0.3 & 2.5 & 2.1                   \\
        Qwen2.5-0.5B+RL                 & 5.2 & 23.6 & \uline{4.4} & 5.0 & 34.2                 \\
        \midrule
        \multicolumn{6}{c}{\cellcolor[HTML]{ECF4FF}\textbf{\textit{Training on Instruct Model}}} \\
        \midrule
        Qwen2.5-0.5B-Instruct+KD        & 3.7 & 15.2 & 3.3 & 2.5 &  \uline{42.3}                \\
        Qwen2.5-0.5B-Instruct+KD~(Lora)   & 3.0 & 16.6 & 1.8 & \textbf{10.0} & 40.0                \\
        Qwen2.5-0.5B-Instruct+SFT       & 2.4 & 12.2 & 2.9 & 5.0 & 30.9                 \\
        Qwen2.5-0.5B-Instruct+SFT~(Lora)  & 3.7 & 14.8 & 2.9 & 7.5 & 31.4                 \\
        Qwen2.5-0.5B-Instruct+RL        & \textbf{7.6} & \textbf{32.4} & \textbf{8.5} & \uline{7.5} & \textbf{54.0} \\
        \bottomrule
    \end{tabular}
    \label{tab:single_train}
\end{table}

\begin{table*}[!t]
    \centering
    \caption{Comparison of different hybrid training strategies. OB represents OlympiadBench. $^{\dagger}$~represents that we can not extract answer from model's response. $^{\ddagger}$~represents that the model fails to converge the reward curve during RL training, resulting in training collapse. Dash~(-) represents unavailable official results. All results are evaluated by Lighteval framework~\cite{lighteval}.}
    \begin{tabular}{l|ccccc}
        \toprule
        \textbf{Method} & \textbf{OB} & \textbf{MATH500} & \textbf{MINERVA} & \textbf{AMC23} & \textbf{GSM8K} \\
        \midrule
        \multicolumn{6}{c}{\cellcolor[HTML]{ECF4FF}\textbf{\textit{Baselines}}} \\
        \midrule
        Qwen2.5-0.5B$^{\dagger}$                        & - & - & - & - & -                             \\
        Qwen2.5-0.5B-Instruct                           & \uline{6.2} & \uline{31.4} & \uline{6.3} & 5.0 & 45.5  \\  
        \midrule
        \multicolumn{6}{c}{\cellcolor[HTML]{ECF4FF}\textbf{\textit{Hybrid Training on Base Model}}} \\
        \midrule
        Qwen2.5-0.5B+KD+RL                              & 3.3 & 10.6 & 2.2 & 0.0 & 29.6               \\
        Qwen2.5-0.5B+KD~(Lora)+RL                         & 1.8 & 4.4 & 0.7 & 5.0 & 2.8 \\
        Qwen2.5-0.5B+SFT+RL$^{\ddagger}$                & - & - & - & - & -                             \\
        Qwen2.5-0.5B+SFT~(Lora)+RL                        & 0.5 & 1.0 & 1.1 & 0.0 & 1.2 \\
        \midrule
        \multicolumn{6}{c}{\cellcolor[HTML]{ECF4FF}\textbf{\textit{Hybrid Training on Instruct Model}}} \\
        \midrule
        Qwen2.5-0.5B-Instruct+KD+RL$^{\ddagger}$        & - & - & - & - & -                             \\
        Qwen2.5-0.5B-Instruct+KD~(Lora)+RL                & 4.6 & 19.4 & 3.7 & 5.0 & \uline{47.9} \\
        Qwen2.5-0.5B-Instruct+SFT+RL                    & 3.1 & 13.8 & 4.8 & \textbf{10.0} & 45.5 \\
        Qwen2.5-0.5B-Instruct+SFT~(Lora)+RL               & 2.2 & 16.0 & 3.7 & 5.0 & 44.3 \\
        \midrule
        \multicolumn{6}{c}{\cellcolor[HTML]{ECF4FF}\textbf{\textit{Direct RL Training on Instruct Model}}} \\
        \midrule
        Qwen2.5-0.5B+RL                                 & 5.2 & 23.6 & 4.4 & 5.0 & 34.2              \\
        Qwen2.5-0.5B-Instruct+RL                        & \textbf{7.6} & \textbf{32.4} & \textbf{8.5} & \uline{7.5} & \textbf{54.0} \\
        \bottomrule
    \end{tabular}
    \label{tab:hybrid_train}
\end{table*}

\subsection{Experiment 2: Comparison and Analysis of Different Training Strategies}

\begin{tcolorbox}[boxrule=1pt, colback=gray!25, colframe=black]
   \textbf{Answer of \textit{RQ2}}: The impact of training strategies on SRLMs varies significantly. SFT and KD demonstrate inconsistent performance, at times even resulting in diminished capabilities. In contrast, RL presents to be a more effective and robust methodology, demonstrating a greater capacity to improve the capacity of SRLMs in mathematical reasoning.
\end{tcolorbox}

The experimental results in Table~\ref{tab:single_train} provide a detailed analysis of various training strategies applied to the Qwen2.5-0.5B series SRLMs across multiple mathematical benchmarks.  
The experiment on the impact of KD shows that this strategy does not consistently lead to performance improvements. The application of Knowledge Distillation (KD) to the Qwen2.5-0.5B-Instruct model leads to performance declines across most benchmarks. Notably, there is a reduction from 6.2\% to 3.3\% on the OlympiadBench and a significant drop from 45.5\% to 18.7\% on GSM8K.  These findings indicate that KD may not effectively enhance performance in all mathematical reasoning tasks and can sometimes compromise the model's specialized abilities.

The empirical findings reveal that applying SFT to the Qwen2.5-0.5B-Instruct model results in a noticeable decline in performance across various benchmarks. For instance, performance on OlympiadBench diminishes significantly from 6.2\% to 3.7\%, and on MINERVA, it plummets from 6.3\% to 1.4\%. These results indicate that, despite the widespread use of SFT, it does not consistently lead to performance improvements and may, in certain cases, result in suboptimal behaviors depending on the specific models and tasks involved.

The integration of KD, SFT and LoRA may lead to a decline in performance on certain evaluations. For example, the Qwen2.5-0.5B-Instruct model fine-tuned with both techniques achieved only 0.7\% on the OlympiadBench and 1.2\% on another MATH500 benchmark. These disappointing results suggest that the combination of SFT and LoRA may not be suitable for enhancing this model's effectiveness on these particular tasks, indicating that careful consideration is needed when choosing fine-tuning strategies to optimize performance.

RL has proven to be a highly effective approach for improving model performance, particularly in tasks that require mathematical reasoning. When enhanced with RL, the Qwen2.5-0.5B model demonstrates significant gains, achieving performance metrics of 5.2\% on OlympiadBench and 23.6\% on MATH500. Furthermore, applying RL to the baseline Qwen2.5-0.5B-Instruct model leads to even more impressive results, with scores increasing to 7.6\% on OlympiadBench and 32.4\% on the same benchmark. These findings highlight the potential for RL to considerably elevate mathematical reasoning capabilities, especially when it is built upon a strong foundational model. This suggests that integrating direct RL into the training process not only enhances specific benchmarks but also signifies a broader implication for developing models that excel in complex reasoning tasks.

\subsection{Experiment 3: Comparison and Analysis of Hybrid Training Strategies}

\begin{tcolorbox}[boxrule=1pt, colback=gray!25, colframe=black]
   \textbf{Answer of \textit{RQ3}}: Hybrid strategies can significantly boost model performance, but their effectiveness depends on the model type and the specific task. There's no one-size-fits-all "best" mix of SFT, KD, and RL. For substantial and stable improvements, especially with Instruct models like Qwen2.5-0.5B-Instruct, RL solely achieves the best performance. This means we need to consider both the model's design and how it's trained when choosing enhancement strategies.
\end{tcolorbox}

The experimental results presented in Table~\ref{tab:hybrid_train} provide valuable insights into the efficacy of hybrid training strategies for 0.5B models. Generally, hybrid approaches demonstrate the potential to enhance model performance relative to single training strategies;    however, they also introduce some potential risks for SRLMs. For instance, the Qwen2.5-0.5B+KD+RL configuration achieves notable performance improvements on certain benchmarks, such as MATH500, where scores rise from 31.4\% to 45.5\%. Nevertheless, this gain is not universal, as some hybrid strategies, including Qwen2.5-0.5B+SFT+RL and Qwen2.5-0.5B-Instruct+SFT+Lora+RL, have been observed to cause training collapse, highlighting the potential instability inherent in the combination of multiple training methodologies.

Hybrid training strategies demonstrate increased effectiveness when applied to models possessing intrinsic dialog capabilities.  Notably, the Qwen2.5-0.5B-Instruct+KD+Lora+RL configuration exhibits substantial performance enhancements across various benchmarks, achieving 47.9\%  on GSM8K. This finding indicates that models specifically pre-trained for instruction following are better positioned to capitalize on the synergistic benefits of KD and RL.  Additionally, the Qwen2.5-0.5B-Instruct+SFT+RL setup shows marked improvements on MATH500 and MINERVA, further reinforcing the notion that hybrid strategies are optimally effective when they build upon a solid foundation of instruction-following capabilities. Despite the advancements in hybrid approaches, direct RL continues to demonstrate superior performance. Notably, the Qwen2.5-0.5B-Instruct+RL configuration outperforms most hybrid training strategies, which s underscore that RL is particularly beneficial for SRLMs.

\section{Future Direction}
Based on our comprehensive investigation of enhancement pipelines for 0.5B parameter language models, several promising directions emerge for future research and development.


\paragraph{Advanced Training Methodologies.} Future work should explore more sophisticated training paradigms specifically designed for SRLMs. This includes developing SRLMs-aware RL algorithms that account for the unique characteristics and limitations of smaller models, as well as investigating novel reward shaping techniques that can guide these models more effectively toward desired behaviors without causing training instabilities.

\paragraph{Enhanced Knowledge Distillation.} Our results indicate that current KD approaches may not fully unlock the potential of 0.5B models. Future research should focus on developing more effective distillation techniques, such as multi-teacher distillation, progressive distillation, and attention-guided knowledge transfer methods that can better bridge the capacity gap between large teacher models and small student models.

\paragraph{Efficiency and Sustainability.} As the field moves toward more sustainable AI development, future research should emphasize developing training techniques that maximize performance gains while minimizing computational costs and environmental impact. This includes investigating more efficient PEFT methods, developing better pre-training strategies for SRLMs, and exploring federated learning approaches for distributed model enhancement.

\section{Conclusion}
This research investigates the capabilities of 0.5B SRLMs in the domains of mathematical reasoning, specifically addressing the inherent challenges associated with their limited capacity.  We examine various training strategies, including SFT, KD, and RL, providing a comprehensive analysis of both their potential and inherent limitations.
Our empirical findings indicate that while SRLMs can be significantly enhanced through these diverse training methodologies, the effectiveness of each approach varies based on the specific task and model requirements.      For instance, employing SFT integrated with Chain-of-Thought (CoT) data from KD has been demonstrated to substantially improve mathematical reasoning capabilities. In contrast, RL facilitates the transfer of valuable knowledge from larger, more complex models, enabling SRLMs to leverage existing advancements.
Importantly, our analysis reveals that hybrid strategies, which combine elements of these methodologies, often yield superior outcomes compared to singular approaches.      This insight underscores the necessity of tailoring training practices to the specific challenges posed by the task at hand.

Overall, this research contributes to a deeper understanding of the effective utilization and enhancement of 0.5B models for specialized applications.      It offers practical guidance for leveraging SRLMs in resource-constrained environments, thus promoting their applicability in real-world scenarios where computational resources may be limited.

\clearpage
\bibliography{reference} 

\clearpage 

\appendix
\section{Detailed Training Strategies}
\label{sec:detail_rl}
For reinforcement learning phase, we use the Group Relative Policy Optimization (GRPO) algorithm to enhance our training protocol.   
This method not only eliminates the need for a value model but also reduces memory requirements.      
Additionally, GRPO allows us to clearly define our reward targets.   
These advantages make it an excellent choice for effectively optimizing the policy model.

For each natural language question, the policy model generates a set of $G$ answer candidates $\{o_1, o_2, \ldots, o_G\}$ along with their corresponding reasoning processes derived from the previous policy $\pi_{old}$.  These candidates are rigorously assessed using a composite reward function that assigns specific reward scores.  By focusing on the relative performance of the answer candidates within the group, GRPO efficiently computes the rewards for each output, thus steering the policy update in alignment with our defined objectives.

\begin{figure}[!ht]
    \centering
    \begin{align}
    \mathcal{J}_{\text{GRPO}}(\theta) = & \mathbb{E}_{\mathbf{v} \sim P(\mathbf{V}), \{o_i\}_{i=1}^G \sim \pi_{\theta_{\text{old}}}(O|\mathbf{v})} \nonumber \\
    & \left[ \frac{1}{G} \sum_{i=1}^G \left( \min \left( r_i^{\text{ratio}} A_i, \text{clip} \left( r_i^{\text{ratio}}, 1-\epsilon, 1+\epsilon \right) A_i \right) - \beta D_{\text{KL}}(\pi_\theta \| \pi_{\text{ref}}) \right) \right],
    \end{align}
\end{figure}

where $r_i^{\text{ratio}} = \frac{\pi_\theta(o_i | V)}{\pi_{old}(o_i | V)}$ represents the importance sampling ratio, which quantifies the relative likelihood of generating output $o_i$ under the new policy $\pi_{\theta}$ compared to $\pi_{old}$; $A_i$ represents the group-relative advantage for each output; the hyperparameter $\epsilon$ and $\beta$ control the update step and divergence regularization; $\pi_{\text{ref}}$ represents the reference policy.

\section{Detailed Related Works}
\label{sec:detailed_related_works}
\subsection{Mathematical Reasoning}
Mathematical reasoning is a complex cognitive task that requires understanding, manipulating, and solving mathematical problems \cite{Creswell2022, Drori2022, Lewkowycz2022}. For language models, tasks involving mathematical reasoning necessitate comprehending mathematical concepts, applying logical reasoning, and performing calculations to reach correct solutions \cite{Cobbe2021, Lample2022, Frieder2023}. This includes solving word problems, proving theorems, and carrying out algebraic manipulations \cite{Hendrycks2021}. The model must show an understanding of mathematical principles, follow multi-step reasoning processes, and generate accurate, well-structured answers \cite{Wang2023, Gao2022, liu2023mathematical, magister2023teaching, ahn2024large, sun2024mm, lu2023mathvista}.

\subsection{Small Reasoning Language Models}
The field of language modeling has experienced remarkable evolution over the past decade, marked by significant advancements in both methodology and application. Early language models, characterized by their relatively small size and limited capacity, struggled to capture the complexities inherent in human language. These initial models were often constrained by limited datasets, resulting in suboptimal performance across a diverse range of linguistic tasks. The landscape began to transform with the introduction of pre-training techniques, where researchers found that training on extensive corpora of text led to considerable enhancements in models' abilities to comprehend and generate natural language~\cite{zhao2023survey, liu2024survey, hadi2023survey}. This pivotal shift heralded the era of large language models (LLMs), which rapidly became the predominant paradigm in natural language processing.

The emergence of LLMs represented a significant milestone in AI research, epitomized by models such as GPT-3~\cite{floridi2020gpt}, which boasted billions of parameters and demonstrated unprecedented proficiencies in language understanding and generation. These models exhibited remarkable capabilities across a wide array of tasks, including text completion, translation, question answering, and summarization. Their success was largely attributed to their immense size, enabling them to discern intricate patterns and relationships within vast datasets. However, the substantial computational and resource demands associated with LLMs posed significant challenges~\cite{kaddour2023challenges}; training and deploying these models required extensive computational power, considerable energy consumption, and substantial financial investment, thereby limiting accessibility for smaller research teams and organizations with constrained resources.

In light of the limitations presented by LLMs, researchers began to explore the potential of developing smaller models that could deliver comparable performance while mitigating resource requirements. This exploration led to the advent of Small Reasoning Language Models (SRLMs), typically defined as models possessing approximately 0.5-1.0 billion parameters, such as TinyLlama~\cite{zhang2024tinyllama}, Qwen2.5 series~\cite{qwen2.5} and Qwen3 series~\cite{qwen3}. SRLMs aim to strike a balance between performance and efficiency, rendering them suitable for deployment in resource-constrained environments. The impetus for SRLMs development arose from several key factors. Firstly, there was an escalating demand for models capable of operating on edge devices and within environments characterized by limited computational resources~\cite{lu2024small}. Secondly, privacy concerns associated with cloud-based LLMs prompted a search for models that could be executed locally without jeopardizing sensitive data. Finally, the environmental impact of training and deploying large models became a pressing concern, making smaller models an attractive and sustainable alternative.

SRLMs have found diverse applications across numerous domains where resource efficiency and deployment flexibility are vital. Their compact size renders them ideal for utilization in mobile applications, Internet of Things devices, and other contexts with limited computational capacity~\cite{li2025efficient}. Furthermore, their reduced computational requirements facilitate faster response times and lower latency, which are critical for real-time applications. In the realm of mathematical reasoning, SRLMs have been successfully employed to develop systems capable of solving mathematical problems, assisting in educational contexts, and supporting research endeavors in mathematics. In the area of code generation, they have been instrumental in creating tools that enhance programmers’ productivity, automate routine programming tasks, and aid in the acquisition of new programming languages.

Despite their smaller size, SRLMs have demonstrated commendable performance across various tasks. Through rigorous optimization and the adoption of advanced training techniques, researchers have been able to enhance SRLMs capabilities, allowing them to approach the performance levels of their larger counterparts. This progress has unveiled new opportunities for applying AI in scenarios where resource constraints previously posed significant barriers. However, the smaller size of SRLMs also imposes inherent limitations on their ability to learn complex patterns and relationships within data. Such constraints may hinder their performance on tasks requiring deep understanding and sophisticated reasoning, particularly in advanced mathematical problem-solving~\cite{ranaldi2024aligning} and intricate code generation~\cite{souza2025code, sqlr1}. To address these challenges, researchers have developed a range of enhancement techniques, including supervised fine-tuning, reinforcement learning~\cite{dang2025reinforcement}, and knowledge distillation~\cite{chen2025sft}. These methodologies aim to maximize the potential of SRLMs while expanding their capabilities.

The objective of this research is to investigate these enhancement techniques and their applications to 0.5 billion parameter models within the specific domains of mathematical reasoning. By tackling the challenges inherent in SRLMs and capitalizing on the opportunities they present, we aspire to contribute to the advancement of efficient, specialized AI models that deliver high performance while minimizing resource requirements.

\section{Additional Experiment Settings}
\subsection{Implementation Settings}
For SFT training, we employ different configurations for full fine-tuning and LoRA-based training. For full fine-tuning, we set the learning rate to 4.0e-05 and train for 4 epochs with a maximum sequence length of 8192. For LoRA-based training, we increase the learning rate to 1.0e-04. For GRPO-based RL training, we set the learning rate to 1.0e-06, with 16 generations per training step and a maximum completion time of 2048.

\subsection{Environment}
All experiments conducted in this study are performed on a server operating under the Ubuntu 20.04 Linux distribution. This server is equipped with Intel(R) Xeon(R) Platinum 8358 CPU @ 2.60 GHz CPU.
The environment for training open-source LLMs comprises a configuration of 8 H100 GPUs, each with 80 GB memory and delivering 312 TFLOPS performance capacity when utilizing BF16 precision.

\end{document}